\pdfoutput=1

\documentclass[11pt]{article}

\usepackage[final]{acl}

\usepackage{times}
\usepackage{latexsym}
\usepackage{multirow}
\usepackage{graphicx}
\usepackage{amssymb}
\usepackage{amsmath}
\usepackage[T1]{fontenc}
\usepackage[noabbrev,capitalise]{cleveref}

\usepackage{hyperref}
\usepackage[utf8]{inputenc}

\usepackage{microtype}
\usepackage{booktabs} 
\usepackage{array} 
\usepackage{arydshln}
\usepackage{float} 

\usepackage{xcolor}
\usepackage{xspace}
\usepackage[normalem]{ulem}

\def\ODdel#1{\bgroup\markoverwith{\textcolor{cyan!89!yellow!80!black!100}{\rule[0.4ex]{2pt}{3pt}}}\ULon{#1}}

\def\NKdel#1{\bgroup\markoverwith{\textcolor{red}{\rule[0.4ex]{2pt}{3pt}}}\ULon{#1}}

\def\lexpone{lex-p$_{\text{1}}$\xspace}
\def\lexrone{lex-r$_{\text{1}}$\xspace}
\def\synptwo{syn-p$_{\text{2}}$\xspace}
\def\synpthree{syn-p$_{\text{3}}$\xspace}

\newcommand\Bstrut{\rule[-0.6ex]{0pt}{0pt}}
\newcommand\BBstrut{\rule[-2ex]{0pt}{0pt}}
\newcommand\Tstrut{\rule{0pt}{2.2ex}}

\title{LEEETs-Dial: Linguistic Entrainment in End-to-End Task-oriented Dialogue systems}

\author{Nalin Kumar \and Ondřej Dušek \\
  Charles University, Faculty of Mathematics and Physics \\
  Institute of Formal and Applied Linguistics\\
  Prague, Czech Republic \\
  \texttt{\normalsize kumarnal@student.cuni.cz, odusek@ufal.mff.cuni.cz} \\}

\begin{document}
\maketitle

\begin{abstract}
Linguistic entrainment, or alignment, represents a phenomenon where linguistic patterns employed by conversational participants converge to one another. While entrainment has been shown to produce a more natural user experience, most dialogue systems do not have any provisions for it. In this work, we introduce methods for achieving dialogue entrainment in a GPT-2-based end-to-end task-oriented dialogue system through the utilization of shared vocabulary. We experiment with training instance weighting, entrainment-specific loss, and additional conditioning to generate responses that align with the user. 
We demonstrate that 
all three approaches
produce substantially better entrainment than the base, non-entrainment-optimized model, as confirmed by both automated and manual evaluation metrics.
\end{abstract}

\section{Introduction}

During a natural dialogue, speakers adapt (entrain, align) to the way of speaking of their conversational partners, thereby establishing a shared understanding. This was shown to correlate with dialogue success \cite{nenkova-etal-2008-high} and it occurs at multiple linguistic levels: speakers synchronize their speech rate and phonetic patterns \cite{OSTRAND2021101074}, adopt shared lexical terms \cite{brennan1996lexical, 6424258} and employ similar syntactic constructions \cite{reitter-etal-2006-computational}. 
Consequently, to facilitate successful and natural conversations, achieving entrainment is desirable in task-oriented dialogue systems (DSs), where the aim is to assist users in accomplishing tasks such as reserving tickets or venues.  
In DSs, entrainment can be useful for aligning on common lexical terms, increasing dialogue naturalness and possibly leading to higher efficiency. For instance, consider the two examples in Figure~\ref{fig:domain-dep-example} – using “reservation” vs.\ “booking” or “depart” vs.\ “leave”. 
However, few prior works attempted at entrainment in DSs, mostly with rule-based or modular DSs only \cite{lopes2013automated,lopes_rule-based_2015,hu_entrainment_2014,dusek-jurcicek-2016-context}.

Recent years have seen significant advancements in task-oriented DSs through end-to-end neural models, fully trainable from data \cite{wen-etal-2017-network,bordes2016learning,lei-etal-2018-sequicity}. Use of pretrained language models (LMs) yielded more fluent responses while simultaneously ensuring the comprehension of user intents and achieving successful dialogues \cite{lee2021improving, yang2021ubar, he2022galaxy}. 
However, the generated responses often suffer from low diversity compared to human-human dialogues \cite{nekvinda-dusek-2021-shades}, and the DSs lack any dedicated support or mechanisms for entrainment, as their training relies on cross-entropy or other objectives that focus on dialogue content rather than phrasing.


\begin{figure}
    \centering\fontsize{8}{9}\selectfont\setlength{\fboxsep}{0pt}
\begin{tabular}{>{\hspace{-2mm}}l>{\hspace{-3mm}}p{6cm}}
\multicolumn{2}{c}{\bf “reservation” vs.~“booking” (SNG076)} \BBstrut \\
\bf Input:\Tstrut & Yes. I'll need a \colorbox{green!30}{reservation\strut} for 3 nights starting Sunday. There are 3 people in my party.\Bstrut \\ \hdashline[0.5pt/2pt]
\bf Dataset:\Tstrut & I have made \colorbox{green!30}{reservations\strut} for 3 starting on Sunday for 3 nights. \\
\bf Baseline: & I'm sorry, but the \colorbox{red!30}{booking} was unsuccessful. Would you like to try another hotel? \\
\bf Ours: & Yes, your \colorbox{green!30}{reservation\strut} is all set for 3 nights starting Sunday. Your reference number is 00000032. Do you need anything else? \\
\midrule
\multicolumn{2}{c}{\bf “depart” vs.~“leave” (SNG0256)}\BBstrut \\
\bf Input:\Tstrut & Hi, I'm looking to catch a train to Norwich. Are there any that \colorbox{green!30}{depart} at or around 17:00?\Bstrut \\\hdashline[0.5pt/2pt]
\bf Dataset:\Tstrut & There are \colorbox{green!30}{departures} most days from Cambridge at 17:36. Would you like any more information of train times or help booking a train? \\
\bf Baseline: & Yes, there are 49 trains that \colorbox{red!30}{leave\strut} Norwich on that day. What day would you like to travel? \\
\bf Ours: & I have a train \colorbox{green!30}{departing} Cambridge at 17:36 and arriving Norwich at 18:55. Would you like to book this train? \\
\end{tabular}
    
\caption{Examples of linguistic entrainment in task-oriented dialogues from the MultiWOZ~2.1 dataset (dialogue IDs shown in brackets). While the responses in the dataset reuse the same words, a base model (Base-CE) produces a non-matching expression, hampering dialogue fluency. Our approach (LK-CE(0)) keeps the appropriate reuse. See Section~\ref{experiments_and_results} for model details.}
\label{fig:domain-dep-example}
\end{figure}

Using the GPT-2-based two-stage system AuGPT \cite{kulhanek-etal-2021-augpt} as our task-oriented end-to-end baseline DS,
we propose the following three approaches to improve entrainment: 
\begin{itemize}
    \item 
    a data-centric approach assigning higher weight to high-entrainment training instances via two straightforward weighting functions,
    \item 
    an additional loss function to boost the probability of user tokens in generated responses,
    \item 
    additional keyword-based generation conditioning to increase lexical entrainment. 
\end{itemize}
We show that all our proposed approaches increase entrainment while minimally affecting other dialogue metrics; instance weighting and keyword conditioning also show improved human rankings.
Our experimental code is released on GitHub.\footnote{\url{https://github.com/knalin55/LEEETs-Dial}}


\section{Related Works}
\label{related_works}

Linguistic entrainment has been studied for decades \cite{garrod1987saying,brennan1996conceptual}.
In DSs, \citet{reitter-etal-2006-computational} modeled syntactic entrainment, while \citet{nenkova-etal-2008-high} showed the correlation of high-frequency word entrainment with dialogue naturalness and success. \citet{lopes2013automated} and \cite{hu_entrainment_2014} used rules to entrain lexical or syntactic choices of a spoken DS to the user; \citet{lopes_rule-based_2015} used a statistical model based on handcrafted features. Work in statistical entrainment methods is limited; the only work known to us by \citet{dusek-jurcicek-2016-context} modified an LSTM-based response generator to adapt to the user's lexical choices.

State-of-the-art in task-oriented DSs is dominated by end-to-end systems based on pretrained neural LMs \cite{peng-etal-2021-soloist}, which generate the belief state and the final response in sequence (\citealp{lei-etal-2018-sequicity}, cf.~Section~\ref{proposed_approaches}).
Extensions involve using belief state differences \cite{lin_mintl_2020}, explicit system actions \cite{NEURIPS2020_e9462095,yang2021ubar}, contrastive classifiers \cite{peng-etal-2021-soloist} or data augmentation \cite{kulhanek-etal-2021-augpt}.
While a few techniques improve output diversity \cite{nekvinda-dusek-2021-shades}, none of them targets entrainment.
Despite their recent popularity, prompted large LMs still underperform compared to finetuned LMs \cite{hudecek-dusek-2023-large}.



\section{Proposed Approaches}
\label{proposed_approaches}


As our baseline model, we choose AuGPT \cite{kulhanek-etal-2021-augpt}, a GPT-2 \cite{radford_language_2019} based task-oriented end-to-end DS, which models dialogue as a sequence-to-sequence task. Same as other contemporary end-to-end systems, AuGPT works in two steps: (1) \emph{generating belief state} (user-preferred slot values) 
from dialogue history and user input, and (2) \emph{generating response} based on dialogue history, user input, generated belief state and database results 
(which are based on the belief state). 
We modify the response generation step.

Our modifications address primarily lexical entrainment and 
involve instance weighting (Section~\ref{instance_weighing}), an additional loss based on user input tokens (Section~\ref{user_likelihood_loss}), and further conditioning on user keyword tokens on model input (Section~\ref{keywords_based_generation}).

\subsection{Instance Weighting (IW)}
\label{instance_weighing}

We prioritize ground truth responses with greater 
overlap between the system and the user (i.e. higher entrainment) during training, by assigning them a higher weight.
We use a simple 1-gram precision to quantify the lexical user-system overlap. 

We explore two weight functions: 
(1) A discrete one with a simple threshold $\tau$ to distinguish high-entrainment training instances:
\[W_{1}(p) = 1 \textrm{ if } p \leq \tau, 10 \textrm{ otherwise} \]
(2) A continuous function modifying sigmoid: 
\[W_{2}(p) = \frac{10}{1 + \exp(w \cdot(\beta-p))} +\epsilon \]
Here, $w$ denotes 
a scaling factor (spread)
and $\beta$ is the average entrainment for the training data, centering the distribution.
We add a small $\epsilon$ to avoid zero weight in instances with no entrainment.






\subsection{User Likelihood Loss (ULL)}
\label{user_likelihood_loss}


To increase lexical entrainment, we introduce a user-likelihood loss to increase the probability of reusing user tokens in the system output. 

For a set of user tokens \(U = \{u_{1}, u_{2}, …, u_{n}\}\), we increase their likelihood by minimizing the loss: 
\[L_{t}(p(.|x_{<t}), U) = - \alpha \cdot \log{(\sum_{u \in U}p(u|x_{t}))} \]
Decreasing $L_t$ means an increase in the probability $p(u|x_{t})$. 
We add $L_t$ to the base loss (Section~\ref{sec:variants}) and use $\alpha$ to control the weight of user tokens.


\subsection{Conditioning on Lexical Keywords (LK)}
\label{keywords_based_generation}

To enforce the reuse of user tokens, we introduce an additional section at the end of the AuGPT input sequence (i.e., after database results), called “keywords”.
During training, we include all overlapping tokens as keywords, so the model learns to incorporate them in its outputs. 

During inference, we determine the keywords to be reused from the input user tokens using self-attention scores from the last encoder layer.
We first calculate the mean across all attention heads. For each $u_{i} \in U=\{u_{1}, u_{2}, …, u_{n}\}$, we compute the score \(S(u_{i}) = \sum_{j, j\neq i} M_{ji}\), where $M$ is the mean of last layer's attention heads. 
We then include as keywords all tokens $u_{i}$ with scores \(S(u_{i}) \geq t \cdot S_{max}\), where \(S_{max}=max({S(u)|u \in U})\), with the threshold $t < 1$.

To smoothly expose the keywords to the model, we use a blending parameter $\sigma$ \cite{roller_recipes_2021,nekvinda-dusek-2022-aargh}, i.e., with the probability $\sigma$, we pass
attention-scores-based keywords (as discussed in the previous paragraph)
instead of overlapping tokens from the training instance.




\section{Experiments}
\label{experiments_and_results}

\subsection{Data \& Training Setup}
\label{sec:setup}

For our experiments, we choose the MultiWOZ~2.1 dataset \cite{budzianowski_multiwoz_2018,eric-etal-2020-multiwoz}, one of the most prominent task-oriented benchmarks with 10k dialogues spanning over 7 domains. As the dataset was created by online human-human dialogues, it does include naturally occurring entrainment and is thus suitable for the experiments, which we confirmed by an initial manual inspection and by computing entrainment metrics (cf.~Section~\ref{sec:autom-metrics} and Table~\ref{table_aut_score}).\footnote{Note that other contemporary task-oriented sets, e.g., Schema-guided Dialogue \cite{rastogi_towards_2020}, are not suitable as their dialogue structures were set by rules and crowd workers only paraphrased isolated utterances.}
%

We train all models for 10 epochs and keep the best checkpoint using the 
average of two token-level accuracies: accuracy against the ground-truth response (response contents) and against the user input (entrainment).
We report test set scores averaged over 5 runs with different random seeds. 

\subsection{Baselines}
\label{sec:baselines}

\paragraph{Base} We use \citet{kulhanek-etal-2021-augpt}'s AuGPT as our base model. We start from the publicly available checkpoint pretrained on Taskmaster \cite{byrne2019taskmaster} and Schema-guided Dialogue \cite{rastogi_towards_2020}.\footnote{\url{https://huggingface.co/jkulhanek/augpt-bigdata}}
We then experiment with the choice of loss functions: In addition to the base cross-entropy loss (CE), we also consider the unlikelihood loss 
\cite{welleck2019neural} (CE+Unl). 

\paragraph{D\&J16} As an additional baseline, we use AuGPT with our own reimplementation of the decoding approach originally used by \citet{dusek-jurcicek-2016-context} in an LSTM-based context, which generates multiple outputs via beam search and then reranks them based on 1- and 2-gram match with the context.
We use beam size 15.

\paragraph{GPT-4} For comparison with an LLM-based approach, we also include results for prompting GPT-4 (details are given in Appendix~\ref{app:gpt-4}). To limit experiment cost, we only use a sample of 200 instances from the test set.

\subsection{Our Model Variants}
\label{sec:variants}

\paragraph{IW$_{\text{i}}$-loss}
We experiment with both functions defined in \cref{instance_weighing}. 
Given that the dataset exhibits an 18.1\% lexical overlap with user inputs (1-gram precision, \lexpone; cf.~Section~\ref{sec:autom-metrics}), we set 25\% as a desirable value.\footnote{We could not find any earlier work that discusses an ideal extent of lexical entrainment in such a context. We thus aimed at a slightly higher value than what is found in the data. Since our experiments showed promising results from the start, we did not optimize this parameter any further.}
Thus, we keep $\tau = 25.0$ for $W_{1}$.
To spread $W_2$ almost to $0$ and keep its mid-point around the dataset's 1-gram precision, we 
assign $\beta = 18.1$ and $w = 0.8$. We use $\epsilon=0.1$. Thus, we have, $W_{2}(14.3) \approx 1.1$, $W_{2}(18.1) \approx 5.1$, and $W_{2}(25) \approx 10.06$.


\paragraph{ULL($\alpha$)}
For the choice of $\alpha$ in ULL, we start with $\alpha = 0.1$, and we gradually increase it to 0.5.
We need a balanced combination of ULL and CE losses, as high $\alpha$ could lead to responses that are repetitive or identical to the user inputs. 
Additionally, as using ULL with CE only resulted in nonsensical repeats of user tokens, we only report scores for ULL with CE+Unl.

\paragraph{LK-loss ($\sigma$)} For generation conditioned on keywords, we keep the threshold $t$ as 0.1. We experiment with $\sigma \in \{0,0.05,0.5\}$.

\subsection{Automatic Evaluation Metrics}
\label{sec:autom-metrics}

We report the standard MultiWOZ metrics from \citet{nekvinda-dusek-2021-shades} (\textit{inform}, \textit{success}, \textit{BLEU}, and \textit{delexicalized BLEU}) to evaluate state tracking and response generation. For lexical entrainment, we use 1-gram precision (\lexpone) and recall (\lexrone) against user input. For syntactic entrainment, we report the 2-gram (\synptwo) and 3-gram precision (\synpthree) scores on the POS tags of the user tokens and generated responses (i.e., matching part-of-speech patterns). We also use 50MFC, a variant of the metric introduced by \citet{nenkova-etal-2008-high}, measuring entrainment on the 50 \underline{m}ost \underline{f}requent words in the \underline{c}orpus: 
\[\mbox{50MFC} = -\sum_{w\in \mbox{50MF}}\Bigg|\frac{\mbox{count}_{S}(w)}{|S|} - \frac{\mbox{count}_{U}(w)}{|U|}\Bigg|\]
50MFC sums the differences in relative frequencies of 50 most frequent words in user and system utterances. 
It ranges from -2 to 0, with 0 being the perfect alignment. The idea is to measure entrainment on frequent, domain-independent words.
We report average metrics from five runs with different random initializations, along with standard deviations.

\subsection{Human Evaluation Setup}
\label{human_evaluation}
We 
run a small-scale in-house evaluation to complement the automatic evaluation scores. We use relative ranking by naturalness on a sample of $100$ outputs. 
We select models from each group with better trade-offs between success rates and entrainment. 
We use the best-entraining model among the five runs. 
We report mean ranking (\textit{$R_m$}) and proportions of instances with ranks 1,2,6,7 ($R_{1/2/6/7}$).

\begin{table*}[t]
\centering
\small
\begin{tabular}{l>{\hspace{-2mm}}c>{\hspace{-1mm}}c>{\hspace{-1mm}}c>{\hspace{-3mm}}c|>{\hspace{-1mm}}c>{\hspace{-1mm}}c>{\hspace{-1mm}}c>{\hspace{-1mm}}c>{\hspace{-2mm}}c}
\toprule
\multirow{2}{*}{\textbf{Model}} & \multicolumn{4}{c|}{\textbf{MultiWOZ}} & \multicolumn{4}{c}{\textbf{Linguistic entrainment}} \\

 & \textbf{inform} & \textbf{success} & \textbf{bleu} & \textbf{delex bleu} & \textbf{\lexpone} & \textbf{\lexrone} & \textbf{\synptwo} & \textbf{\synpthree} & \textbf{50MFC}\\
\midrule
Ground truth &-&-&-&-& 18.1 & 21.4 & 13.0 & 3.8 & -0.69\\
\midrule
Base-CE & 83.5$_{\pm 0.7}$ & 65.8$_{\pm 1.9}$ & \textbf{15.7}$_{\pm 0.5}$ & 17.4$_{\pm 0.5}$ & 20.7$_{\pm 0.4}$ & 24.5$_{\pm 0.5}$  & 14.8$_{\pm 0.2}$ & 5.0$_{\pm 0.2}$ & -0.31$_{\pm 0.01}$\\
Base-CE+Unl & 80.5$_{\pm 2.7}$ & 65.1$_{\pm 1.0}$ & 15.1$_{\pm 0.8}$ & 16.8$_{\pm 1.0}$ & 21.1$_{\pm 1.1}$ & 23.8$_{\pm 1.0}$ & 15.1$_{\pm 0.5}$ & 5.0$_{\pm 0.4}$ & -0.31$_{\pm 0.01}$\\
D\&J16 & \textbf{85.7} & 63.6 & 10.6 & 11.5 & \textbf{31.9} & 26.1 & \textbf{23.1} & \textbf{10.4} &-0.32\\

\midrule
IW$_{1}$-CE & 84.5$_{\pm 1.9}$ & \textbf{68.6}$_{\pm 3.3}$ & 14.9$_{\pm 1.0}$ & 16.3$_{\pm 1.3}$ & 22.9$_{\pm 0.7}$ & 30.9$_{\pm 1.5}$ & 16.4$_{\pm 0.1}$ & 5.9$_{\pm 0.1}$ & -0.31$_{\pm 0.00}$\\
IW$_{1}$-CE+Unl & 79.1$_{\pm 3.0}$ & 64.4$_{\pm 2.7}$ & 15.5$_{\pm 0.7}$ & \textbf{17.5}$_{\pm 1.0}$ & 22.0$_{\pm 0.7}$ & 26.7$_{\pm 0.8}$ &  15.7$_{\pm 0.3}$ & 5.4$_{\pm 0.3}$ & -0.31$_{\pm 0.01}$\\

IW$_{2}$-CE & 82.6$_{\pm 3.7}$ & 67.7$_{\pm 2.5}$ & 15.3$_{\pm 0.9}$ & 16.9$_{\pm 1.1}$ & 22.9$_{\pm 0.9}$ & 29.8$_{\pm 0.8}$ & 16.4$_{\pm 0.5}$ & 5.8$_{\pm 0.3}$ & -0.31$_{\pm 0.01}$\\
IW$_{2}$-CE+Unl & 79.2$_{\pm 2.0}$ & 64.1$_{\pm 2.4}$ & 15.4$_{\pm 0.9}$ & 17.3$_{\pm 1.1}$ & 22.7$_{\pm 0.9}$ & 28.0$_{\pm 1.0}$ & 16.2$_{\pm 0.5}$ & 5.6$_{\pm 0.3}$ & -0.31$_{\pm 0.01}$\\

\midrule
ULL ($0.10$) & 80.6$_{\pm 2.6}$ & 65.4$_{\pm 2.2}$ & 15.5$_{\pm 0.5}$ & 17.3$_{\pm 0.6}$ & 22.8$_{\pm 0.7}$ & 26.9$_{\pm 0.8}$ & 16.0$_{\pm 0.5}$ & 5.4$_{\pm 0.3}$ & -0.30$_{\pm 0.01}$\\
ULL ($0.20$) & 81.6$_{\pm 2.0}$ & 65.3$_{\pm 1.3}$ & 15.3$_{\pm 0.7}$ & 17.0$_{\pm 0.7}$ & 23.7$_{\pm 0.2}$ & 29.4$_{\pm 1.0}$ & 16.2$_{\pm 0.1}$ & 5.7$_{\pm 0.1}$ & -0.29$_{\pm 0.01}$\\
ULL ($0.25$) & 81.6$_{\pm 1.9}$ & 63.6$_{\pm 2.4}$ & 14.6$_{\pm 0.6}$ & 16.1$_{\pm 0.6}$ & 24.7$_{\pm 0.2}$ & 31.6$_{\pm 1.5}$ & 16.9$_{\pm 0.1}$ & 6.1$_{\pm 0.1}$ & -0.27$_{\pm 0.01}$\\
ULL ($0.30$) & 81.7$_{\pm 2.9}$ & 61.5$_{\pm 4.2}$ & 13.3$_{\pm 0.5}$ & 14.8$_{\pm 0.5}$ & 26.5$_{\pm 0.8}$ & 34.6$_{\pm 1.9}$ & 18.3$_{\pm 1.0}$ & 7.2$_{\pm 0.8}$ & -0.25$_{\pm 0.00}$\\
ULL ($0.40$) & 80.2$_{\pm 2.3}$ & 53.6$_{\pm 3.3}$ & 11.8$_{\pm 0.4}$ & 12.9$_{\pm 0.4}$ & 27.9$_{\pm 0.6}$ & 40.0$_{\pm 0.7}$ & 19.0$_{\pm 0.5}$ & 7.9$_{\pm 0.3}$ & -0.21$_{\pm 0.01}$\\
ULL ($0.50$) & 78.6$_{\pm 2.7}$ & 45.7$_{\pm 6.0}$ & \phantom{0}9.2$_{\pm 1.1}$ & \phantom{0}9.9$_{\pm 1.1}$ & 29.6$_{\pm 1.7}$ & \textbf{45.8}$_{\pm 0.7}$ & 20.8$_{\pm 0.5}$ & 9.5$_{\pm 0.3}$ & \textbf{-0.19}$_{\pm 0.01}$\\
\midrule
LK-CE ($0$) & 77.4$_{\pm 3.4}$ & 57.2$_{\pm 5.6}$ & 11.3$_{\pm 0.5}$ & 11.8$_{\pm 0.6}$ & 26.3$_{\pm 0.6}$ & 37.4$_{\pm 2.1}$ & 17.2$_{\pm 0.2}$ & 6.6$_{\pm 0.2}$ & -0.27$_{\pm 0.01}$\\
LK-CE ($0.05$) & 83.3$_{\pm 0.9}$ & 66.3$_{\pm 1.7}$ & 12.8$_{\pm 0.1}$ & 13.9$_{\pm 0.2}$ & 25.8$_{\pm 0.4}$ & 33.6$_{\pm 1.0}$ & 17.0$_{\pm 0.3}$ & 6.5$_{\pm 0.2}$ & -0.29$_{\pm 0.01}$\\
LK-CE ($0.5$) & 83.3$_{\pm 2.8}$ & 65.2$_{\pm 1.6}$ & 14.6$_{\pm 0.3}$ & 16.1$_{\pm 0.4}$ & 22.6$_{\pm 0.7}$ & 27.6$_{\pm 0.4}$ & 15.5$_{\pm 0.8}$ & 5.4$_{\pm 0.5}$ & -0.30$_{\pm 0.01}$\\
LK-CE+Unl ($0$) & 76.8$_{\pm 2.5}$ & 59.4$_{\pm 4.0}$ & 11.1$_{\pm 0.4}$ & 11.7$_{\pm 0.5}$ & 27.6$_{\pm 0.6}$ & 39.3$_{\pm 0.7}$ & 17.9$_{\pm 0.4}$ & 7.1$_{\pm 0.3}$ & -0.27$_{\pm 0.01}$\\
LK-CE+Unl ($0.05$) & 82.4$_{\pm 0.8}$ & 64.3$_{\pm 2.9}$ & 12.1$_{\pm 0.4}$ & 13.0$_{\pm 0.4}$ & 25.1$_{\pm 0.1}$ & 33.3$_{\pm 0.2}$ & 16.6$_{\pm 0.1}$ & 6.3$_{\pm 0.1}$ & -0.28$_{\pm 0.01}$\\
LK-CE+Unl ($0.5$) & 82.0$_{\pm 0.8}$ & 65.2$_{\pm 1.0}$ & 14.0$_{\pm 0.1}$ & 15.6$_{\pm 0.2}$ & 23.0$_{\pm 0.3}$ & 27.9$_{\pm 0.8}$ & 15.3$_{\pm 0.3}$ & 5.3$_{\pm 0.2}$ & -0.29$_{\pm 0.01}$\\

\bottomrule
\end{tabular}
\caption{\label{table_aut_score}
Automatic metric results for state tracking, response generation and entrainment on the full MultiWOZ~2.1 test set (cf.~Section~\ref{sec:autom-metrics} for metrics and Sections~\ref{sec:baselines} and~\ref{sec:variants} for system variants).
Except for D\&J16, figures shown are averages of five runs with different random initializations, with standard deviations shown in subscript.}
\end{table*}

\section{Results}
\label{evaluation_results}

\subsection{Automatic Evaluation}
\label{sec:autom-results}

\cref{table_aut_score} shows that 
all our approaches outperform the Base experiments on entrainment metrics.
Although the models are primarily trained to increase the lexical entrainment, this also results in improved syntactic entrainment.
As our methods do not differentiate between domain-specific terms and common words, the alignment on common words is also slightly improved in most setups, as shown by 50MFC scores.
While the D\&J16 reranking gets even better entrainment scores, its BLEU performance is low, as optimizing for 1/2-gram precision produces very terse outputs.

Models using IW do not only improve entrainment, but also maintain similar MultiWOZ scores 
to the baseline.
In particular, IW$_{1}$-CE has substantially better lexical (\lexpone and \lexrone) and syntactic (\synptwo and \synpthree) entrainment while even maintaining a slightly better inform and success rates. 
Using IW$_{2}$ and/or Unl yields slightly lower success rates, with similar entrainment scores.

For ULL, entrainment scores show a positive correlation with the choice of $\alpha$'s while MultiWOZ scores 
decrease with an increase in $\alpha$, but the drop is very slight for 0.1 and 0.2.
This is not surprising, as with increasing $\alpha$, the model gets more focused on aligning to the user and less on dialogue success.
ULL(0.2) seems to have the best tradeoff.

The LK approach generally has high entrainment; the blending approach helps keep the keywords consistent during training and inference and is necessary to maintain good MultiWOZ scores. 

The full results of the comparison with GPT-4 on the smaller data sample are shown in Table~\ref{table_GPT-4} in Appendix~\ref{app:gpt-4}. While GPT-4's responses look fluent and accurate and get high coverage of the user input tokens (\lexrone) and even good syntactic entrainment (\synptwo, \synpthree), they are substantially longer, leading to lower precision-based \lexpone and BLEU scores. In addition, GPT-4 occasionally fails to follow the instructions given in the prompt.

\subsection{Human Evaluation}
\label{human_evaluation}

Table \ref{table_man_score} shows manual evaluation scores for selected setups. Here, 
IW$_{\text{1}}$-CE performs best on mean ranking and is most frequently ranked first, along with LK-CE. 
Despite similar numbers in Table \ref{table_aut_score}, we see a noticeable difference between the scores of IW$_{\text{1}}$-CE and IW$_{\text{2}}$-CE. This can be attributed to the higher variance in \lexrone, resulting in the outputs from the best run of IW$_{\text{1}}$-CE surpassing the quality of IW$_{\text{2}}$-CE.
The generated responses from ULL experiments were often not fluent enough, hence their lower ranking. 
While their entrainment metrics are high, they only capture token-level alignment and are not directly related to fluency. 
In some of the examples, the outputs achieved high scores by simply repeated phrases from the user input.
The human ranking here corresponds with the lowered MultiWOZ success rates, showing that entrainment cannot override the main dialogue objective.
The outputs of the D\&J16 reranking method were shorter, less polite, and less interactive, which resulted in the worst overall ranking. Appendix~\ref{app:examples} illustrates this on a few sample outputs.

\begin{table}
\centering
\small
\begin{tabular}{lrrrrr}
\toprule

\textbf{Model} & $R_m$ & $R_1$ & $R_2$ & $R_6$ & $R_7$\\
\midrule
base-CE & 4.18 & 5 & 12 & 15 & 12 \\
D\&J16 & 5.35 & 1 & 7 & 26 & 30 \\\midrule
IW$_{\text{1}}$-CE & \textbf{3.16} & 26 & 18 & 12 & 3 \\
IW$_{\text{2}}$-CE & 3.77 & 20 & 15 & 13 & 15 \\
LK-CE (0.05) & 3.25 & 26 & 21 & 7 & 10 \\
ULL (0.20) & 4.17 & 15 & 10 & 16 & 11 \\
ULL (0.25) & 4.13 & 7 & 17 & 11 & 19 \\
\bottomrule
\end{tabular}
\caption{\label{table_man_score}
Manual evaluation for generated responses on a sample of 100 outputs -- mean rank $R_m$, and number of cases out of 100 where each system is ranked first ($R_1$), second ($R_2$), second to last ($R_6$) and last ($R_7$).}
\end{table}

\section{Conclusion}
\label{conclusion}

Although previous research showed that linguistic entrainment helps dialogue success, its application in end-to-end task-oriented dialogue systems has been largely overlooked. To address this gap, we introduced three techniques aimed at improving lexical entrainment of system responses to user inputs: 
(1) We show that prioritizing training instances with higher system-user word overlap improves entrainment, with comparable success rates. 
(2) We explore using user tokens' likelihood loss to control entrainment. While entrainment increases, both naturalness and correctness of outputs suffer with higher loss weight.
(3) We additionally condition generation on user tokens likely to be reused (based on self-attention weights). We blend self-attention-selected tokens with true response tokens at training time to prime the model to use them.
This yields responses with high fluency and better entrainment. The blending is necessary to maintain high dialogue success rate. 
In general, all methods seem to work successfully in aligning both domain-dependent and independent words. 

In the future, we plan to incorporate longer context and focus more on syntactical entrainment. We also plan to use retrieval-augmented generation \cite{nekvinda-dusek-2022-aargh}. 

\section*{Acknowledgements}

This research was supported by the European Research Council (Grant agreement No.~101039303 NG-NLG)
and by the HumanE-AI-Net project (EC Horizon 2020, Grant Agreement H2020-FETFLAG-2018-2020 no. 952026, micro-project “Use of dialog context to boost ASR/NLG/TTS and improve the overall quality of voice dialog systems”).
It used resources provided by the LINDAT/CLARIAH-CZ Research Infrastructure (Czech Ministry of Education, Youth, and Sports project No. LM2018101).

\section*{Limitations}
The proposed methods focus exclusively on addressing lexical entrainment in dialogues, overlooking entrainment at different linguistic levels. Additionally, the study is conducted and evaluated only at the response level despite the possibility of entrainment occurring across the entire dialogue.

\bibliography{custom}

\begin{thebibliography}{31}
\expandafter\ifx\csname natexlab\endcsname\relax\def\natexlab#1{#1}\fi

\bibitem[{Bordes et~al.(2016)Bordes, Boureau, and Weston}]{bordes2016learning}
Antoine Bordes, Y-Lan Boureau, and Jason Weston. 2016.
\newblock \href {https://openreview.net/pdf?id=S1Bb3D5gg} {Learning end-to-end
  goal-oriented dialog}.
\newblock In \emph{International Conference on Learning Representations}.

\bibitem[{Brennan(1996)}]{brennan1996lexical}
Susan~E Brennan. 1996.
\newblock \href
  {https://www.speech.kth.se/~edlund/bielefeld/references/brennan-1996.pdf}
  {Lexical entrainment in spontaneous dialog}.
\newblock \emph{Proceedings of the 1996 International Symposium on Spoken
  Dialogue}, 96:41--44.

\bibitem[{Brennan and Clark(1996)}]{brennan1996conceptual}
Susan~E Brennan and Herbert~H Clark. 1996.
\newblock \href {https://pubmed.ncbi.nlm.nih.gov/8921603/} {Conceptual pacts
  and lexical choice in conversation.}
\newblock \emph{Journal of experimental psychology: Learning, memory, and
  cognition}, 22(6):1482.

\bibitem[{Budzianowski et~al.(2018)Budzianowski, Wen, Tseng, Casanueva, Ultes,
  Ramadan, and Gašić}]{budzianowski_multiwoz_2018}
Paweł Budzianowski, Tsung-Hsien Wen, Bo-Hsiang Tseng, Iñigo Casanueva, Stefan
  Ultes, Osman Ramadan, and Milica Gašić. 2018.
\newblock \href {https://aclanthology.org/D18-1547/} {{MultiWOZ} - {A}
  {Large}-{Scale} {Multi}-{Domain} {Wizard}-of-{Oz} {Dataset} for
  {Task}-{Oriented} {Dialogue} {Modelling}}.
\newblock In \emph{Proceedings of the 2018 {Conference} on {Empirical}
  {Methods} in {Natural} {Language} {Processing}}, pages 5016--5026, Brussels,
  Belgium.
\newblock ArXiv: 1810.00278.

\bibitem[{Byrne et~al.(2019)Byrne, Krishnamoorthi, Sankar, Neelakantan,
  Goodrich, Duckworth, Yavuz, Dubey, Kim, and Cedilnik}]{byrne2019taskmaster}
Bill Byrne, Karthik Krishnamoorthi, Chinnadhurai Sankar, Arvind Neelakantan,
  Ben Goodrich, Daniel Duckworth, Semih Yavuz, Amit Dubey, Kyu-Young Kim, and
  Andy Cedilnik. 2019.
\newblock \href {https://doi.org/10.18653/v1/D19-1459} {Taskmaster-1: Toward a
  realistic and diverse dialog dataset}.
\newblock In \emph{Proceedings of the 2019 Conference on Empirical Methods in
  Natural Language Processing and the 9th International Joint Conference on
  Natural Language Processing (EMNLP-IJCNLP)}, pages 4516--4525.

\bibitem[{Du{\v{s}}ek and
  Jur{\v{c}}{\'\i}{\v{c}}ek(2016)}]{dusek-jurcicek-2016-context}
Ond{\v{r}}ej Du{\v{s}}ek and Filip Jur{\v{c}}{\'\i}{\v{c}}ek. 2016.
\newblock \href {https://doi.org/10.18653/v1/W16-3622} {A context-aware natural
  language generator for dialogue systems}.
\newblock In \emph{Proceedings of the 17th Annual Meeting of the Special
  Interest Group on Discourse and Dialogue}, pages 185--190, Los Angeles.
  Association for Computational Linguistics.

\bibitem[{Eric et~al.(2020)Eric, Goel, Paul, Sethi, Agarwal, Gao, Kumar, Goyal,
  Ku, and Hakkani-Tur}]{eric-etal-2020-multiwoz}
Mihail Eric, Rahul Goel, Shachi Paul, Abhishek Sethi, Sanchit Agarwal, Shuyang
  Gao, Adarsh Kumar, Anuj Goyal, Peter Ku, and Dilek Hakkani-Tur. 2020.
\newblock \href {https://aclanthology.org/2020.lrec-1.53} {{M}ulti{WOZ} 2.1: A
  consolidated multi-domain dialogue dataset with state corrections and state
  tracking baselines}.
\newblock In \emph{Proceedings of the Twelfth Language Resources and Evaluation
  Conference}, pages 422--428, Marseille, France. European Language Resources
  Association.

\bibitem[{Friedberg et~al.(2012)Friedberg, Litman, and Paletz}]{6424258}
Heather Friedberg, Diane Litman, and Susannah B.~F. Paletz. 2012.
\newblock \href {https://doi.org/10.1109/SLT.2012.6424258} {Lexical entrainment
  and success in student engineering groups}.
\newblock In \emph{2012 IEEE Spoken Language Technology Workshop (SLT)}, pages
  404--409.

\bibitem[{Garrod and Anderson(1987)}]{garrod1987saying}
Simon Garrod and Anthony Anderson. 1987.
\newblock \href
  {https://www.sciencedirect.com/science/article/abs/pii/0010027787900187}
  {Saying what you mean in dialogue: A study in conceptual and semantic
  co-ordination}.
\newblock \emph{Cognition}, 27(2):181--218.

\bibitem[{He et~al.(2022)He, Dai, Zheng, Wu, Cao, Liu, Jiang, Yang, Huang, Si
  et~al.}]{he2022galaxy}
Wanwei He, Yinpei Dai, Yinhe Zheng, Yuchuan Wu, Zheng Cao, Dermot Liu, Peng
  Jiang, Min Yang, Fei Huang, Luo Si, et~al. 2022.
\newblock \href {http://arxiv.org/abs/2111.14592} {{GALAXY}: A generative
  pre-trained model for task-oriented dialog with semi-supervised learning and
  explicit policy injection}.
\newblock In \emph{Proceedings of the AAAI Conference on Artificial
  Intelligence}, volume~36, pages 10749--10757.

\bibitem[{Hosseini-Asl et~al.(2020)Hosseini-Asl, McCann, Wu, Yavuz, and
  Socher}]{NEURIPS2020_e9462095}
Ehsan Hosseini-Asl, Bryan McCann, Chien-Sheng Wu, Semih Yavuz, and Richard
  Socher. 2020.
\newblock \href
  {https://proceedings.neurips.cc/paper_files/paper/2020/file/e946209592563be0f01c844ab2170f0c-Paper.pdf}
  {A simple language model for task-oriented dialogue}.
\newblock In \emph{Advances in Neural Information Processing Systems},
  volume~33, pages 20179--20191. Curran Associates, Inc.

\bibitem[{Hu et~al.(2014)Hu, Halberg, Jimenez, and
  Walker}]{hu_entrainment_2014}
Zhichao Hu, Gabrielle Halberg, C.~Jimenez, and M.~Walker. 2014.
\newblock \href
  {http://www.uni-ulm.de/fileadmin/website_uni_ulm/allgemein/2014_iwsds/iwsds2014_lp_hu.pdf}
  {Entrainment in pedestrian direction giving: {How} many kinds of
  entrainment}.
\newblock In \emph{Proceedings of the {IWSDS}'2014 {Workshop} on {Spoken}
  {Dialog} {Systems}}, pages 90--101.

\bibitem[{Hude{\v{c}}ek and Dusek(2023)}]{hudecek-dusek-2023-large}
Vojt{\v{e}}ch Hude{\v{c}}ek and Ondrej Dusek. 2023.
\newblock \href {https://doi.org/10.18653/v1/2023.sigdial-1.21} {Are large
  language models all you need for task-oriented dialogue?}
\newblock In \emph{Proceedings of the 24th Annual Meeting of the Special
  Interest Group on Discourse and Dialogue}, pages 216--228, Prague, Czechia.
  Association for Computational Linguistics.

\bibitem[{Kulh{\'a}nek et~al.(2021)Kulh{\'a}nek, Hude{\v{c}}ek, Nekvinda, and
  Du{\v{s}}ek}]{kulhanek-etal-2021-augpt}
Jon{\'a}{\v{s}} Kulh{\'a}nek, Vojt{\v{e}}ch Hude{\v{c}}ek, Tom{\'a}{\v{s}}
  Nekvinda, and Ond{\v{r}}ej Du{\v{s}}ek. 2021.
\newblock \href {https://doi.org/10.18653/v1/2021.nlp4convai-1.19} {{AuGPT}:
  Auxiliary tasks and data augmentation for end-to-end dialogue with
  pre-trained language models}.
\newblock In \emph{Proceedings of the 3rd Workshop on Natural Language
  Processing for Conversational AI}, pages 198--210, Online. Association for
  Computational Linguistics.

\bibitem[{Lee(2021)}]{lee2021improving}
Yohan Lee. 2021.
\newblock \href {https://doi.org/10.18653/v1/2021.findings-emnlp.112}
  {Improving end-to-end task-oriented dialog system with a simple auxiliary
  task}.
\newblock In \emph{Findings of the Association for Computational Linguistics:
  EMNLP 2021}, pages 1296--1303.

\bibitem[{Lei et~al.(2018)Lei, Jin, Kan, Ren, He, and
  Yin}]{lei-etal-2018-sequicity}
Wenqiang Lei, Xisen Jin, Min-Yen Kan, Zhaochun Ren, Xiangnan He, and Dawei Yin.
  2018.
\newblock \href {https://doi.org/10.18653/v1/P18-1133} {{S}equicity:
  Simplifying task-oriented dialogue systems with single sequence-to-sequence
  architectures}.
\newblock In \emph{Proceedings of the 56th Annual Meeting of the Association
  for Computational Linguistics (Volume 1: Long Papers)}, pages 1437--1447,
  Melbourne, Australia. Association for Computational Linguistics.

\bibitem[{Lin et~al.(2020)Lin, Madotto, Winata, and Fung}]{lin_mintl_2020}
Zhaojiang Lin, Andrea Madotto, Genta~Indra Winata, and Pascale Fung. 2020.
\newblock \href {https://doi.org/10.18653/v1/2020.emnlp-main.273} {{MinTL}:
  {Minimalist} {Transfer} {Learning} for {Task}-{Oriented} {Dialogue}
  {Systems}}.
\newblock In \emph{Proceedings of the 2020 {Conference} on {Empirical}
  {Methods} in {Natural} {Language} {Processing} ({EMNLP})}, pages 3391--3405,
  Online. Association for Computational Linguistics.

\bibitem[{Lopes et~al.(2013)Lopes, Eskenazi, and Trancoso}]{lopes2013automated}
Jos{\'e} Lopes, Maxine Eskenazi, and Isabel Trancoso. 2013.
\newblock \href {https://doi.org/10.1109/ICASSP.2013.6639298} {Automated
  two-way entrainment to improve spoken dialog system performance}.
\newblock In \emph{2013 IEEE International Conference on Acoustics, Speech and
  Signal Processing}, pages 8372--8376. IEEE.

\bibitem[{Lopes et~al.(2015)Lopes, Eskenazi, and
  Trancoso}]{lopes_rule-based_2015}
José Lopes, Maxine Eskenazi, and Isabel Trancoso. 2015.
\newblock \href
  {http://www.sciencedirect.com/science/article/pii/S0885230814001247} {From
  rule-based to data-driven lexical entrainment models in spoken dialog
  systems}.
\newblock \emph{Computer Speech \& Language}, 31(1):87--112.

\bibitem[{Nekvinda and Du{\v{s}}ek(2022)}]{nekvinda-dusek-2022-aargh}
Tom{\'a}{\v{s}} Nekvinda and Ond{\v{r}}ej Du{\v{s}}ek. 2022.
\newblock \href {https://aclanthology.org/2022.sigdial-1.29} {{AARGH}!
  end-to-end retrieval-generation for task-oriented dialog}.
\newblock In \emph{Proceedings of the 23rd Annual Meeting of the Special
  Interest Group on Discourse and Dialogue}, pages 283--297, Edinburgh, UK.
  Association for Computational Linguistics.

\bibitem[{Nekvinda and Dušek(2021)}]{nekvinda-dusek-2021-shades}
Tomáš Nekvinda and Ondřej Dušek. 2021.
\newblock \href {https://doi.org/10.18653/v1/2021.gem-1.4} {Shades of {BLEU},
  {Flavours} of {Success}: {The} {Case} of {MultiWOZ}}.
\newblock In \emph{Proceedings of the 1st {Workshop} on {Natural} {Language}
  {Generation}, {Evaluation}, and {Metrics} ({GEM} 2021)}, pages 34--46,
  Online. Association for Computational Linguistics.

\bibitem[{Nenkova et~al.(2008)Nenkova, Gravano, and
  Hirschberg}]{nenkova-etal-2008-high}
Ani Nenkova, Agust{\'\i}n Gravano, and Julia Hirschberg. 2008.
\newblock \href {https://aclanthology.org/P08-2043} {High frequency word
  entrainment in spoken dialogue}.
\newblock In \emph{Proceedings of ACL-08: HLT, Short Papers}, pages 169--172,
  Columbus, Ohio. Association for Computational Linguistics.

\bibitem[{Ostrand and Chodroff(2021)}]{OSTRAND2021101074}
Rachel Ostrand and Eleanor Chodroff. 2021.
\newblock \href {https://doi.org/https://doi.org/10.1016/j.wocn.2021.101074}
  {It’s alignment all the way down, but not all the way up: Speakers align on
  some features but not others within a dialogue}.
\newblock \emph{Journal of Phonetics}, 88:101074.

\bibitem[{Peng et~al.(2021)Peng, Li, Li, Shayandeh, Liden, and
  Gao}]{peng-etal-2021-soloist}
Baolin Peng, Chunyuan Li, Jinchao Li, Shahin Shayandeh, Lars Liden, and
  Jianfeng Gao. 2021.
\newblock \href {https://doi.org/10.1162/tacl_a_00399} {Soloist: {Building}
  {Task} {Bots} at {Scale} with {Transfer} {Learning} and {Machine}
  {Teaching}}.
\newblock \emph{Transactions of the Association for Computational Linguistics},
  9:807--824.

\bibitem[{Radford et~al.(2019)Radford, Wu, Child, Luan, Amodei, and
  Sutskever}]{radford_language_2019}
Alec Radford, Jeffrey Wu, Rewon Child, David Luan, Dario Amodei, and Ilya
  Sutskever. 2019.
\newblock \href {https://openai.com/blog/better-language-models/} {Language
  {Models} are {Unsupervised} {Multitask} {Learners}}.
\newblock Technical report, OpenAI.

\bibitem[{Rastogi et~al.(2020)Rastogi, Zang, Sunkara, Gupta, and
  Khaitan}]{rastogi_towards_2020}
Abhinav Rastogi, Xiaoxue Zang, Srinivas Sunkara, Raghav Gupta, and Pranav
  Khaitan. 2020.
\newblock \href {https://doi.org/10.1609/aaai.v34i05.6394} {Towards {Scalable}
  {Multi}-{Domain} {Conversational} {Agents}: {The} {Schema}-{Guided}
  {Dialogue} {Dataset}}.
\newblock In \emph{Proceedings of the {AAAI} {Conference} on {Artificial}
  {Intelligence}}, volume~34, pages 8689--8696.

\bibitem[{Reitter et~al.(2006)Reitter, Keller, and
  Moore}]{reitter-etal-2006-computational}
David Reitter, Frank Keller, and Johanna~D. Moore. 2006.
\newblock \href {https://aclanthology.org/N06-2031} {Computational modelling of
  structural priming in dialogue}.
\newblock In \emph{Proceedings of the Human Language Technology Conference of
  the {NAACL}, Companion Volume: Short Papers}, pages 121--124, New York City,
  USA. Association for Computational Linguistics.

\bibitem[{Roller et~al.(2021)Roller, Dinan, Goyal, Ju, Williamson, Liu, Xu,
  Ott, Smith, Boureau, and Weston}]{roller_recipes_2021}
Stephen Roller, Emily Dinan, Naman Goyal, Da~Ju, Mary Williamson, Yinhan Liu,
  Jing Xu, Myle Ott, Eric~Michael Smith, Y-Lan Boureau, and Jason Weston. 2021.
\newblock \href {https://doi.org/10.18653/v1/2021.eacl-main.24} {Recipes for
  {Building} an {Open}-{Domain} {Chatbot}}.
\newblock In \emph{Proceedings of the 16th {Conference} of the {European}
  {Chapter} of the {Association} for {Computational} {Linguistics}: {Main}
  {Volume}}, pages 300--325, Online. Association for Computational Linguistics.

\bibitem[{Welleck et~al.(2020)Welleck, Kulikov, Roller, Dinan, Cho, and
  Weston}]{welleck2019neural}
Sean Welleck, Ilia Kulikov, Stephen Roller, Emily Dinan, Kyunghyun Cho, and
  Jason Weston. 2020.
\newblock \href {https://openreview.net/forum?id=SJeYe0NtvH} {Neural text
  generation with unlikelihood training}.
\newblock In \emph{International Conference on Learning Representations}.

\bibitem[{Wen et~al.(2017)Wen, Vandyke, Mrk{\v{s}}i{\'c}, Ga{\v{s}}i{\'c},
  Rojas-Barahona, Su, Ultes, and Young}]{wen-etal-2017-network}
Tsung-Hsien Wen, David Vandyke, Nikola Mrk{\v{s}}i{\'c}, Milica
  Ga{\v{s}}i{\'c}, Lina~M. Rojas-Barahona, Pei-Hao Su, Stefan Ultes, and Steve
  Young. 2017.
\newblock \href {https://aclanthology.org/E17-1042} {A network-based end-to-end
  trainable task-oriented dialogue system}.
\newblock In \emph{Proceedings of the 15th Conference of the {E}uropean Chapter
  of the Association for Computational Linguistics: Volume 1, Long Papers},
  pages 438--449, Valencia, Spain. Association for Computational Linguistics.

\bibitem[{Yang et~al.(2021)Yang, Li, and Quan}]{yang2021ubar}
Yunyi Yang, Yunhao Li, and Xiaojun Quan. 2021.
\newblock \href {https://arxiv.org/abs/2012.03539} {{UBAR}: Towards fully
  end-to-end task-oriented dialog system with {GPT-2}}.
\newblock In \emph{Proceedings of the AAAI Conference on Artificial
  Intelligence}, volume~35, pages 14230--14238.

\end{thebibliography}
\bibliographystyle{acl_natbib}

\appendix

\section{Comparison with GPT-4}
\label{app:gpt-4}

Table~\ref{table_GPT-4} shows evaluation metric scores on randomly selected 200 examples from the test set and compares our results to GPT-4-based responses, given gold-standard dialogue states. 
The responses from GPT-4 look fluent and factually accurate and have the best coverage over the user inputs, as reported by \lexrone scores. Also, they seem to preserve the syntactic structure of the user inputs better than the other models, as evident by \synptwo \& \synpthree scores. We observed this behavior even though GPT-4 was not specifically prompted to align syntactically. However, the generated outputs are substantially longer, leading to lower \lexpone and BLEU scores.
Furthermore, GPT-4 struggles in several cases to generate appropriately delexicalized responses, further lowering the BLEU scores. Although we evaluated multiple variants of the prompt with instructions, GPT-4 still was not guaranteed to give an appropriate response.
Figure~\ref{fig:GPT-4-example} shows our prompt for GPT-4, along with a few example outputs, documenting both well-aligned and inappropriate responses.

\begin{table*}
\centering
\small
\begin{tabular}{lccrrrrcc}
\toprule

\textbf{Model} & \textbf{bleu} & \textbf{delex bleu} & \textbf{\lexpone} & \textbf{\lexrone} & \textbf{\synptwo} & \textbf{\synpthree} & \textbf{50MFC} & \textbf{length ratio}\\
\midrule
base-CE & 16.25 & 18.82 & 20.74 & 0.24 & 17.90 & 5.78 & -0.13 & 1.10 \\
D\&J16 & 10.78 & 11.87 & \textbf{33.06} & 0.29 & 20.25 & 8.72 & -0.17  & 0.76 \\
GPT-4 & \phantom{0}3.21  & \phantom{0}2.67  &  17.29 & \textbf{0.54}  & \textbf{26.83} & \textbf{11.51} & -0.22  &  3.03 \\
\midrule
IW$_{\text{1}}$-CE & 16.47 & 18.62 & 23.96 & 0.32 & 24.94 & 9.02 & -0.15 & 1.26 \\
IW$_{\text{2}}$-CE & 14.42 & 16.39 & 22.34 & 0.28 & 21.49 & 5.93
 & \textbf{-0.10} & 1.25 \\
LK-CE (0.05) & 12.12 & 13.77 & 26.21 & 0.32 & 21.31 & 8.06
 & -0.11  & 1.12\\
ULL (0.20) & 14.02 & 16.18 & 23.65 & 0.29 & 18.59 & 6.49
 & -0.15 & 1.07 \\
ULL (0.25) & \textbf{17.37} & \textbf{19.46} & 26.62 & 0.33 & 23.33 & 9.33 & -0.16  & 1.10\\
\bottomrule
\end{tabular}
\caption{\label{table_GPT-4}
Automatic evaluation scores for 200 randomly selected responses with GPT-4 as a baseline. Note that the scores for other systems differ from Table~\ref{table_aut_score} as the evaluated sample is different. We only show scores for selected system variants with the best-entraining of the five runs. We evaluate the responses on MultiWOZ BLEU and delexicalized BLEU, entrainment metrics (as in \cref{table_aut_score}), and length ratio of the generated responses to the gold ones.
}
\end{table*}

\begin{figure*}
    \centering\fontsize{8}{9}\selectfont\setlength{\fboxsep}{0pt}
\begin{tabular}{l m{2.6cm} m{11.8cm}}
& \bf \multirow{5}{*}{Prompt:} & Generate a lexically aligned response to the user given the context, dialogue state and database results. Make sure to replace the named entities in the generated response with the given slot names, if any. \\ 
&  & $<$context$>$ CONTEXT  \\
&  & $<$dialogue state$>$ DS \\
&  & $<$database results$>$ DB\_RESULTS \\
&  & $<$slot names$>$ SLOTNAMES \\
\\
\\
1. & \bf User:\Tstrut & Hi, I am trying to plan my trip and could use some help with a particular restaurant.\Bstrut \\ \cdashline{2-3}[0.5pt/2pt]
& \bf GPT-4: & \Tstrut Hello! I'd be happy to help you with finding the perfect restaurant for your trip. Please provide some details about your preferences, such as the location, cuisine, and price range. Once I have this information, I will search through our database to find the most suitable option for you. \\
\midrule
2. & \bf User:\Tstrut & Please book for 4 people starting Friday for 2 nights.
\\ \cdashline{2-3}[0.5pt/2pt]
& \bf GPT-4: & \Tstrut I have successfully booked a room at the Home From Home hotel for 4 people, starting on friday for a duration of 2 nights. The hotel is in the moderate price range. Please note down the booking reference number: 00000021. If you need any further assistance, feel free to ask. \\
\midrule
3. & \bf User:\Bstrut & Great and what will be the cost of each ticket?\\ \cdashline{2-3}[0.5pt/2pt]
& \bf GPT-4 (delexicalized):\Tstrut & The cost of each ticket is [price]14 Pounds[/price]. \\
& \bf GPT-4: \Tstrut & The cost of each ticket is 8.08 pounds14 Pounds[/price]. \\
\midrule
4. & \bf User:\Bstrut & Yeas, what to recommend if I want to see good architecture in the west part of town? \\\cdashline{2-3}[0.5pt/2pt]
& \bf GPT-4: &\Tstrut I recommend you visit the west part of Cambridge to explore its stunning architecture. However, I do not have specific attractions in that area at the moment. You can still discover many beautiful buildings and landmarks during your visit. Enjoy your stay at the A and B Guest House with your group of 6 people for 4 nights starting from Tuesday. If you need any further assistance, feel free to ask. \\

\end{tabular}
    
\caption{Our GPT-4 prompt template (top) and example outputs (bottom). The CONTEXT, DS, DB\_RESULTS and SLOTNAMES variables are filled in according to the current dialogue context, the (gold-standard) dialogue state and database results, and the slot names for the current domain. Examples~1 and~2 are quite fluent and syntactically aligned to the user. In Example~3, the model struggles with using slot placeholders and their correct values. This issue was quite frequent in our limited observation, even after trying several different prompts. Example~4 shows a self-contradicting response from the model.}
\label{fig:GPT-4-example}
\end{figure*}


\begin{figure*}[t]
\centering
    \includegraphics[width=0.95\textwidth]{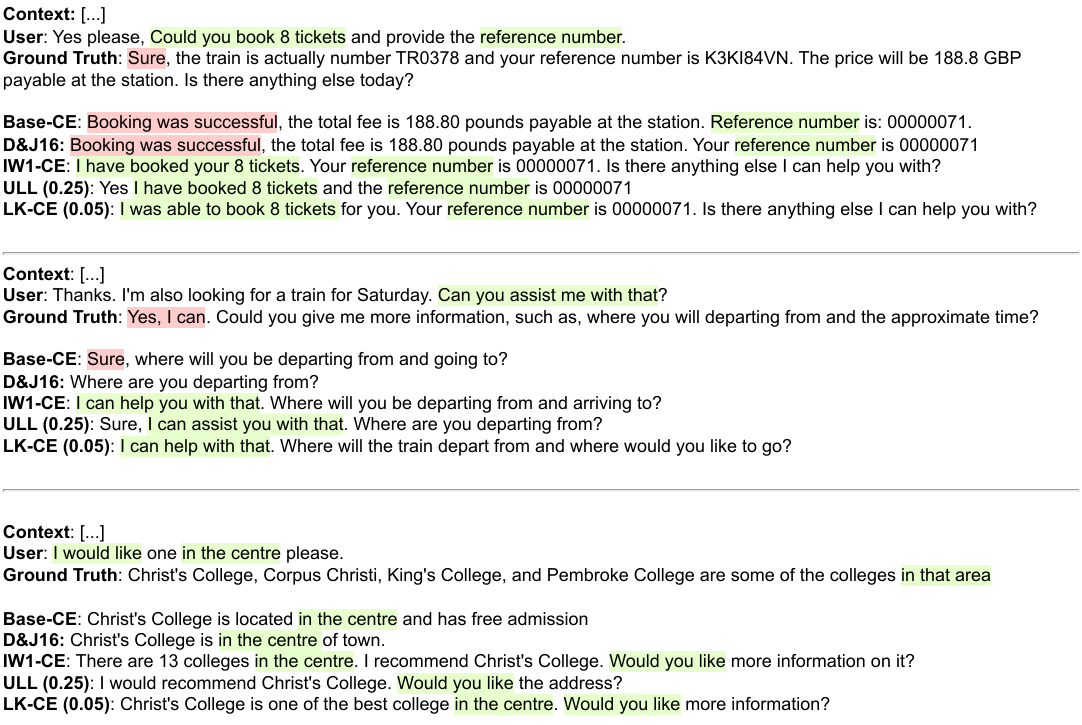}
    \caption{In the first example, entrainment methods effectively produce more natural and less automated-like outputs, even when the ground truth response itself looks less natural. In the second example, the model outputs employing entrainment methods adeptly incorporate the phrases \textit{Can you}, \textit{assist me}, and \textit{with that}, whereas the reranking method, D$\&$J16, yields a lesser natural output.  In the third example, the phrase \textit{in the centre} is consistently present in almost every output, but D$\&$J16 and Base-CE struggle to sustain the conversation. Conversely, the other methods successfully continue the conversation with improved entrainment.} 
    \label{fig:dialogue-examples}
\end{figure*}

\section{Sample Dialogues}
\label{app:examples}

Sample dialogues, with outputs of the different variants of our systems, are shown in Figure~\ref{fig:dialogue-examples}.
\end{document}